\pdfoutput=1
\documentclass[sigconf]{acmart}
\usepackage{color}
\usepackage{bbm}
\usepackage{multirow}
\usepackage[inline]{enumitem}
\usepackage{subfigure} 
\usepackage{sistyle}
\usepackage[ruled]{algorithm2e}
\usepackage{booktabs}
\usepackage{caption}
\SIthousandsep{,}
\usepackage{makecell}
\usepackage[skip=0pt]{caption}
\usepackage{amsmath}
\usepackage{hyperref}
\usepackage{array} 
\usepackage{graphicx}
\usepackage{xcolor}
\usepackage{acronym}
\usepackage[export]{adjustbox}

\newcommand{\heading}[1]{\vspace*{0.5mm}\noindent\textbf{#1.}}

\AtBeginDocument{%
  \providecommand\BibTeX{{%
    \normalfont B\kern-0.5em{\scshape i\kern-0.25em b}\kern-0.8em\TeX}}}

\makeatletter
\g@addto@macro\normalsize{%
  \abovedisplayskip 2pt plus1pt 
  \belowdisplayskip 2pt plus1pt
  \abovedisplayshortskip  2pt plus1pt%
  \belowdisplayshortskip  1pt plus1pt
}
\makeatother

\newcommand{\benchmark}{SpecUBench\xspace} 
\acrodef{IR}{information retrieval}
\acrodef{LLM}{large language model}

\AtBeginDocument{%
  \providecommand\BibTeX{{%
    Bib\TeX}}}

\copyrightyear{2026}
\acmYear{2026}
\setcopyright{cc}
\setcctype{by}
\acmConference[CIKM '26]{Proceedings of the 35th ACM International Conference on Information and Knowledge Management}{November 07--11, 2026}{Rome, Italy}
\acmBooktitle{Proceedings of the 35th ACM International Conference on Information and Knowledge Management (CIKM '26), November 07--11, 2026, Rome, Italy}
\acmDOI{10.1145/3799682.3840666}
\acmISBN{979-8-4007-2539-5/2026/11}
\makeatletter
\gdef\@copyrightpermission{
  \begin{minipage}{0.3\columnwidth}
   \href{https://creativecommons.org/licenses/by/4.0/}{\includegraphics[width=0.90\textwidth]{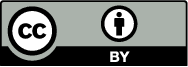}}
  \end{minipage}\hfill
  \begin{minipage}{0.7\columnwidth}
   \href{https://creativecommons.org/licenses/by/4.0/}{This work is licensed under a Creative Commons Attribution International 4.0 License.}
  \end{minipage}
  \vspace{5pt}
}
\author{Hengran Zhang}
\orcid{0009-0004-1144-1298}
\affiliation{
    \institution{State Key Laboratory of AI Safety}
	\institution{Institute of Computing Technology, Chinese Academy of Sciences}
	\institution{University of Chinese Academy of Sciences}
	\city{Beijing}
	\country{China}
}
\email{zhanghengran22z@ict.ac.cn}

\author{Keping Bi}
\orcid{0000-0001-5123-4999}
\authornote{Keping Bi and Jiafeng Guo are the corresponding authors.}
\affiliation{
   	\institution{State Key Laboratory of AI Safety}
	\institution{Institute of Computing Technology, Chinese Academy of Sciences}
	\institution{University of Chinese Academy of Sciences}
	\city{Beijing}
	\country{China}
}
\email{bikeping@ict.ac.cn}

\author{Jiafeng Guo}
\orcid{0000-0002-9509-8674}
\authornotemark[1]
\affiliation{
    \institution{State Key Laboratory of AI Safety}
	\institution{Institute of Computing Technology, Chinese Academy of Sciences}
	\institution{University of Chinese Academy of Sciences}
	\city{Beijing}
	\country{China}
}
\email{guojiafeng@ict.ac.cn}

\author{Jiaming Zhang}
\orcid{0009-0009-8687-4713}
\author{Shuaiqiang Wang}
\orcid{0000-0002-9212-1947}
\author{Dawei Yin}
\orcid{0000-0002-0684-6205}
\affiliation{
	\institution{Baidu Inc}
 \city{Beijing}
 \country{China}
}
\email{zhangjiaming04@baidu.com}
\email{wangshuaiqiang@baidu.com}
\email{yindawei@acm.org}

\author{Xueqi Cheng}
\orcid{0000-0002-5201-8195}
\affiliation{
		\institution{State Key Laboratory of AI Safety}
	\institution{Institute of Computing Technology, Chinese Academy of Sciences}
	\institution{University of Chinese Academy of Sciences}
	\city{Beijing}
	\country{China}
}
\email{cxq@ict.ac.cn}

\renewcommand{\shortauthors}{Hengran Zhang, Keping Bi, Jiafeng Guo, Jiaming Zhang, Shuaiqiang Wang, Dawei Yin, Xueqi Cheng}
\begin{document}

\title{LLM-Specific Utility for Retrieval-Augmented Generation}

\begin{abstract}
Retrieval-augmented generation (RAG) is typically optimized for topical relevance, yet its success ultimately depends on whether retrieved passages are useful for a large language model (LLM) to generate correct and complete answers. We argue that such utility is often LLM-specific rather than universal, due to differences in models’ knowledge, reasoning, and ability to leverage evidence. We formalize LLM-specific utility as the performance improvement of a target LLM when a passage is provided, compared to answering without evidence. To systematically study LLM-specific utility, we construct a benchmark of LLM-specific gold utilitarian passages for four LLMs (Qwen3-8B/14B/32B and Llama 3.1-8B) on six QA datasets (Natural Questions, TriviaQA, MS MARCO-FQA, HotpotQA, FEVER and 2WikiQA). Our analysis shows that utilitarian passages are model-dependent and non-transferable: each LLM performs best with its own utilitarian evidence, while evidence optimized for other LLMs is consistently suboptimal. Human-annotated evidence remains a strong general baseline but does not fully match individual LLM utility needs. We further introduce the LLM-specific utility judgment task and construct the corresponding benchmark, i.e., \textbf{\benchmark} (LLM-Specific Utility Benchmark). Experiments show that existing utility-aware selection and scoring methods largely capture model-agnostic usefulness and struggle to reliably estimate LLM-specific utility. Overall, our findings highlight the limitations of current utility-aware retrieval and motivate generator-tailored evidence selection for improving RAG. 
Our code and datasets can be found at \url{https://github.com/Trustworthy-Information-Access/LLM_specific_utility}.

\end{abstract}


\maketitle
\acresetall

\section{Introduction}
Retrieval-augmented generation (RAG) enhances large language models (LLMs) by grounding generation in external evidence. While retrieval is traditionally optimized for topical relevance—i.e., whether a passage matches the query in content \cite{saracevic1988study, schamber1988relevance}—RAG ultimately succeeds or fails based on a different criterion: whether the retrieved evidence is \textit{useful} for the LLM to produce an accurate and comprehensive answer \cite{zhang2024large, zhang2024iterative}. In practice, passages that appear relevant to humans may not be effectively leveraged by an LLM, and conversely, passages that are not ranked highly by relevance-driven retrieval may still substantially improve generation \cite{zhang2025distilling}. This exposes a fundamental gap between classic retrieval objectives and downstream LLM generation objectives.

This gap can be naturally understood through the lens of \textbf{utility}, a concept that has long been emphasized in web search and interactive information retrieval \cite{saracevic1996relevance, saracevic1975relevance, saracevic1988study, mao2016does, yilmaz2014relevance}. In web search, the goal is not merely to return relevant documents, but to provide results that help users accomplish their underlying intent—such as learning, deciding, buying, troubleshooting, or navigating \cite{mao2016does, machmouchi2017beyond, dai2021beyond}. In the LLM era, however, the \textit{consumer} of retrieved results shifts: instead of directly serving end users, retrieval evidence is primarily consumed by the LLM, which synthesizes information into a final response. As a result, the optimization target shifts from ``utility to users'' toward \textbf{utility to LLMs}, where evidence quality should be judged by whether it improves the LLM’s downstream generation.

\begin{figure}[!t]
    \centering
     \includegraphics[width=\linewidth]{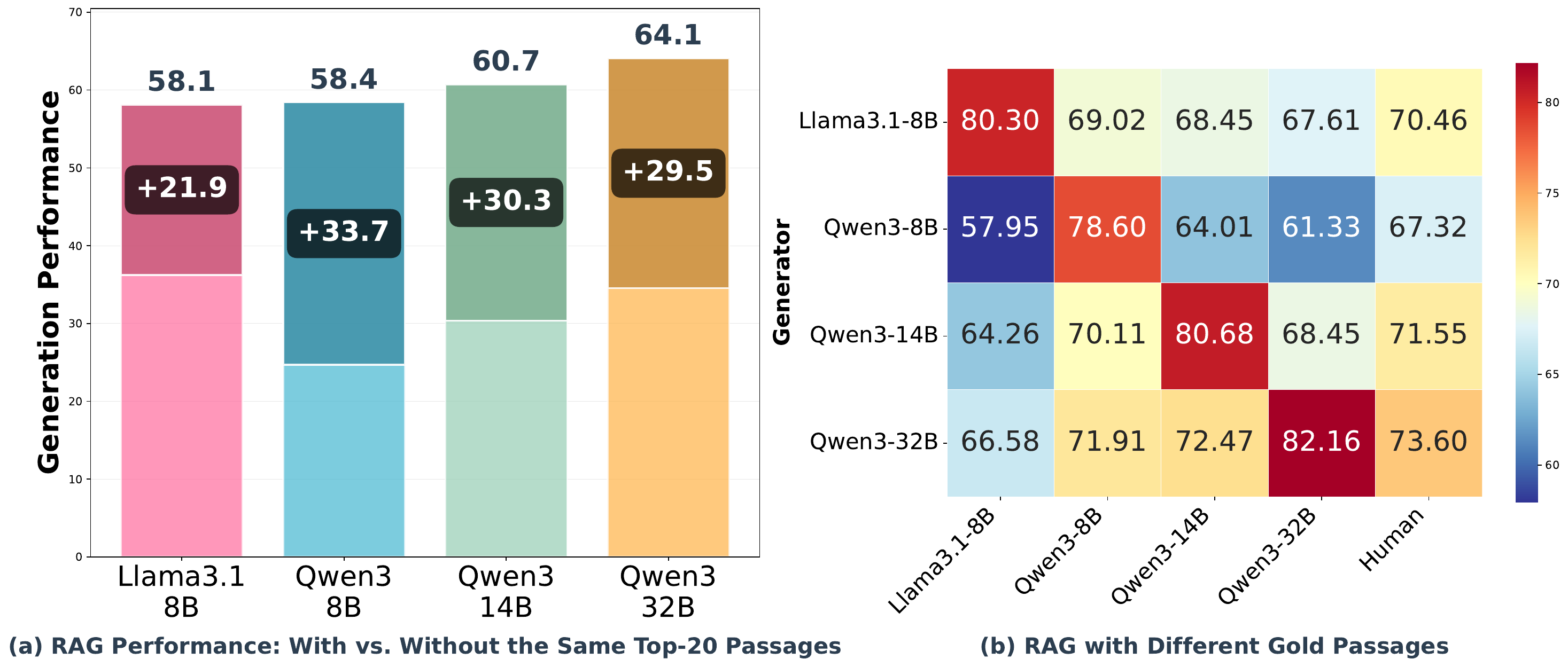}
    \caption{(a) RAG accuracy (\%) with the same top-20 retrieval results upon different LLMs on the NQ dataset. (b) RAG accuracy (\%) of LLMs with gold utilitarian passages for different LLMs on the NQ dataset.}
    \label{fig:utility_relevance}
\end{figure}

Motivated by this perspective, recent RAG research has increasingly moved beyond relevance and explored utility-aware retrieval and selection \cite{zhang2024large, shi2023replug, izacard2023atlas}. Existing approaches typically obtain passage utility signals by measuring generation performance given a passage—e.g., the likelihood of producing the ground-truth answer \cite{izacard2023atlas}, or matching-based metrics such as EM/F1 \cite{zamani2024stochastic}—then using these signals to train or distill retrievers. Another line of work performs post-retrieval selection, prompting LLMs to explicitly judge which passages are useful (often with pseudo-answer generation) \cite{zhang2024large}, using iterative selection frameworks \cite{zhang2024iterative}, or training lightweight selectors that can process more candidates efficiently \cite{zhang2025distilling}. Despite their effectiveness, most prior work implicitly assumes that utility is a \textit{generic} property: a passage deemed utilitarian should benefit any LLM similarly, and thus a single universal notion of ``good evidence'' suffices.

In this work, we challenge this assumption and argue that utility in RAG is often \textbf{LLM-specific}. Similar to personalized web search, where the usefulness of identical search results can differ across users due to intent, expertise, constraints, and preferences, different LLMs may have different utility standards due to differences in training data, internal knowledge, reasoning behavior, and comprehension ability. Concretely, as shown in Figure\ref{fig:utility_relevance}(a), even when the same top-20 retrieval results are provided, different LLMs can exhibit substantially different downstream QA accuracy boost, suggesting that the same retrieved evidence may not be equally beneficial across models. 

To formalize this phenomenon, we define \textbf{LLM-specific utility} as follows: a passage is \textit{utilitarian} for a target LLM if providing it as evidence improves the LLM’s answer generation performance compared to answering without any evidence. 
Under this definition, a utilitarian passage must (i) supply information that fills the model’s \textit{knowledge gap} for the query, and (ii) be \textit{leveragable} by the model during generation. Crucially, a passage can be ``gold'' from a human perspective, i.e., high-quality and relevant evidence, yet still fail to benefit a specific LLM, either because the model already knows the answer or because it cannot reliably interpret, integrate, or apply the passage to produce the correct response.


Based on this principle, we construct LLM-specific gold utilitarian passages and use them to systematically study passage utility from a model-centric perspective. Specifically, we annotate gold utilitarian passages for four LLMs with different scales and architectures (Qwen3-8B, Qwen3-14B, Qwen3-32B~\cite{qwen3technicalreport}, and Llama~3.1-8B~\cite{dubey2024llama}) on six QA datasets: Natural Questions (NQ)~\cite{kwiatkowski2019natural}, TriviaQA~\cite{joshi2017triviaqa}, MS MARCO-FQA (derived from MS MARCO QA~\cite{nguyen2016ms}), HotpotQA \cite{yang2018hotpotqa}, FEVER \cite{thorne2018fever}, and 2WikiQA \cite{xanh2020_2wikimultihop}, using multiple candidate pools to capture diverse evidence sources. Based on this, we primarily study the following research questions:

\textbf{RQ1}: \textit{Do LLMs have model-specific utilitarian passages, and how are they related to human-annotated evidence?}
While prior work typically treats utility as universal, we ask whether utilitarian passages differ across LLMs and whether human gold evidence remains optimal under an LLM-centric definition.

\textbf{RQ2}: \textit{How accurately can existing methods assess LLM-specific utility?}
Given that many utility-aware RAG methods rely on LLM-based judgments or generation-based proxies, we investigate whether these methods truly capture model-specific utility or only recover a generic notion of usefulness.

For \textbf{RQ1}, we make three observations. First, \textit{LLM-specific utilitarian passages exist and are not transferable}: each LLM performs best with its own gold utilitarian passages, while passages optimized for other models are suboptimal. Second, \textit{human-annotated evidence is not optimal but remains a strong general-utility baseline}: it is often the second-best choice, indicating robust yet model-agnostic utility. Figure~\ref{fig:utility_relevance}(b) summarizes these trends by comparing RAG performance using (i) each model’s own gold utilitarian passages, (ii) other models’ gold utilitarian passages, and (iii) human gold passages. Third, the remaining gap is partly due to \textit{LLM comprehension and readability}: even correct evidence may provide little utility if the model cannot effectively leverage it.


To answer \textbf{RQ2}, we introduce the LLM-specific utility judgment task: given a query and retrieved candidate passages, the goal is to identify the subset of passages that are utilitarian for a target LLM.  
We build a benchmark, i.e., \textbf{\benchmark} (LLM-Specific Utility Benchmark), with LLM-specific ground truth and evaluate representative utility judgment methods, including verbalized selection/ranking with pseudo-answers \cite{zhang2024large, zhang2025distilling} and probabilistic utility estimation based on generation likelihood \cite{shi2023replug} and attention weights \cite{izacard2023atlas}. 
Results show that verbalized methods with pseudo-answers achieve better performance on both LLM-specific utility judgment evaluation and downstream RAG performance evaluation compared to other methods. 
However, our findings reveal a critical limitation: when an LLM generates answers using passages selected by its own utility judgments, its downstream performance is not optimal. 
Instead, we observe a universal trend across all evaluated models: an LLM consistently achieves best generation performance when using passage sets selected by a stronger LLM. 
This phenomenon demonstrates that existing methods fail to identify LLM-specific utility. These findings highlight an important limitation of current utility judgment methods and motivate the need for approaches that can capture genuine LLM-specific utility.

Our contributions are summarized as: 
(1) We propose LLM-specific utility as a model-centric retrieval objective for RAG, and construct a benchmark, i.e., \benchmark, with LLM-specific gold utilitarian passages for multiple LLMs across standard QA datasets. 
(2) Through extensive analysis, we show that utilitarian passages are model-dependent and non-transferable, and that human-annotated gold evidence, while strong, does not fully align with the utility needs of individual LLMs. 
(3) We formalize the LLM-specific utility judgment task, and demonstrate that existing utility-aware selection and ranking methods largely capture model-agnostic utility rather than true LLM-specific utility, motivating future work on generator-tailored evidence selection for improving RAG performance.

\section{Related Work}

\subsection{Retrieval-Augmented Generation (RAG)}
Retrieval-augmented generation (RAG) has attracted growing interest as a strategy for mitigating large language models’ hallucinations and knowledge recency/staleness by leveraging retrieved information as external knowledge \cite{lewis2020retrieval, gao2023retrieval, jiang2023active, su2025parametric}.
The current challenges in retrieval-augmented generation (RAG) primarily center on three aspects: 
\textit{how to formulate the information needs of LLMs} \cite{das2019multi, khot2022decomposed, khattab2022demonstrate, shao2023enhancing, trivedi2022interleaving, press2022measuring, yao2023react, jiang2023active, li2023llatrieval}—that is, during the RAG process, enabling LLMs to formalize their information requirements based on available context and express them as natural language queries to retrieve relevant documents. For example, FLARE \cite{press2022measuring} proposed two methods for information need verbalization of LLMs: generate a query for low-confidence spans in the already generated sentence via LLM and masked token with low-confidence as implicit queries.
LLatrieval \cite{li2023llatrieval} prompted LLMs to generate the missing information for answering the question. 
\textit{how to fulfill these information needs} \cite{zhang2024iterative,zhang2024large, zhang2025distilling, zhang2023relevance, zhao2024longrag}, mainly by improving the retrieval of more relevant documents through enhancements to the retriever itself or by selecting useful documents from retrieval results in the post-retrieval stage to better accomplish the task.
For example, UtiSel \cite{zhang2025leveraging} distilled the utility judgments from LLMs to a dense retriever, reducing dependence on costly human annotations and improving retrieval performance with curriculum learning. Further, UtilityQwen \cite{zhang2025distilling} distilled the utility judgments from LLMs to a utility-based selector.
and 
\textit{how to enhance the harmlessness, faithfulness, or helpfulness of the generation process of large language models} \cite{wang2022self, wang2022self, guo2025deepseek}. 

Our work focuses on the second domain—fulfilling LLMs' information needs through evidence selection. Unlike approaches that iteratively reformulate queries, we directly identify evidence that fills each LLM's knowledge gaps from retrieved documents. This aligns with utility-focused RAG literature but explicitly considers \textit{LLM-specific} utility—different LLMs benefit differently from the same evidence due to variations in their internal knowledge and comprehension abilities.

\subsection{Utility-focused RAG}
Traditional retrieval models are trained and evaluated using human-annotated relevance labels \cite{nguyen2016ms, zhang2023relevance}. In RAG, however, the emphasis shifts from optimizing retrieval metrics to maximizing downstream generation performance through retrieval that provides genuinely useful evidence \cite{zhang2025distilling}. 
To improve the utility of retrieved information, three main approaches have emerged: 
(1) Verbalized utility judgments \cite{zhang2024large,zhang2024iterative, zhang2025distilling}: Prompt the LLM to explicitly assess the usefulness of candidate documents, often aided by pseudo-answer generation and iterative refinement. 
To improve the utility judgments performance, \citet{zhang2024iterative} proposes the ITEM framework inspired by the relevance in philosophy using iterative interaction between different components in RAG. 
\citet{zhang2025leveraging} further proposes using automatic LLM annotation of utility to train retrievers with a focus on utility. 
(2) Downstream-performance-based utility estimation \cite{shi2023replug, izacard2023atlas, salemi2025learning, zamani2024stochastic,zhang2023relevance}: Score documents by their impact on end-task performance, such as QA accuracy or the likelihood of producing the ground-truth answer. 
\citet{shi2023replug} employed the likelihood of the ground-truth answer as the utility score of the input passage and then as the guide to distill the retriever. 
For fact verification, \citet{zhang2023relevance} used the loss divergence of ground-truth evidence and the input evidence as the utility estimation to optimize the evidence retrieval process. 
(3) Attention-based utility estimation \cite{izacard2020distilling}: Estimate document utility from the attention mass the generator allocates to tokens from each input document during answer generation, aggregated to the document level. 
While methods like answer likelihood or attention weights capture aspects of LLM-specific behavior, prior work treats utility as model-agnostic—trained retrievers or selectors are deployed generically across different LLMs without explicit consideration of model-specific utility. The key distinction of our work is that we \textit{explicitly study} whether utility is LLM-specific: we construct LLM-specific ground truth, demonstrate that utilitarian passages are model-dependent and non-transferable, and show that existing methods largely capture generic utility rather than genuine LLM-specific utility.

\subsection{3H Principle in LLM Alignment}
After being fine-tuned, large language models (LLMs) acquire the ability to follow instructions and engage in dialogue, enabling preliminary interactions with users. 
Further advancing this line of research, scholars have sought to align LLMs with human values, proposing that LLM outputs should adhere to the 3H principles \cite{wang2023aligning, liu2023trustworthy}: Helpfulness (providing accurate and task-aligned responses), Honesty (avoiding hallucinations and misinformation), and Harmlessness (preventing toxic or unethical outputs). These principles, grounded in human-centric criteria, are enforced through Reinforcement Learning from Human Feedback (RLHF) \cite{touvron2023llama, guo2025deepseek, wang2024arithmetic} to align LLMs'  behavior with human values. 
Unlike helpfulness in 3H principles, this work investigates the utility of retrieval results for specific LLMs, adopting a LLM-centric perspective on utility.


\section{LLM-Specific Utility} 
Existing research on passage utility in RAG often views utility as an inherent property of a document, i.e., whether a passage contains information that would help answer a question, regardless of who or what consumes it.
However, in practice, the ultimate consumer of retrieved evidence in RAG is not the human, but the LLM itself. Crucially, the presence of useful information in a passage does not guarantee that the LLM will be able to interpret, integrate, and apply that information to improve its answer generation. This disconnect motivates a shift from evaluating generic, intrinsic utility to assessing LLM-specific utility. 

\subsection{Formal Definition}
\label{sec:gold_annotation}
Formally, we define the utility of a passage $d$ for a given LLM $\mathcal{L}$ on query $q$ as follows:
\begin{equation}
\label{eq:utility-def}
u(d; \mathcal{L}, q) =
Acc(\mathcal{L}(q, d)) - Acc(\mathcal{L}(q, \emptyset)),
\end{equation}
where $\mathcal{L}(q, \emptyset)$ denotes the response generated by $\mathcal{L}$ without any passage context; $\mathcal{L}(q, d)$ denotes the response generated by $\mathcal{L}$ when provided with passage $d$; and $Acc(\cdot)$ measures the accuracy of the generated answer. Following previous work~\cite{kwiatkowski2019natural, ni2025towards}, we use accuracy to evaluate QA performance, determined by checking whether any gold answer appears as a substring in the model's response after normalization. 

A passage $d$ is deemed \textit{utilitarian} for LLM $\mathcal{L}$ on query $q$ if its utility is positive, i.e., $u(d; \mathcal{L}, q) > 0$. The gold utilitarian passages for LLM $\mathcal{L}$ on query $q$ are then defined as:
\begin{equation}
\label{eq:utilitarian-set}
\mathcal{G}_{q}^{(\mathcal{L})}
=
\{d \in C \mid u(d; \mathcal{L}, q) > 0\},
\end{equation}
where $C$ denotes the candidate pool of passages.

\noindent\subsection{Implications of the Definition}
This formulation captures the essence of LLM-specific utility: a passage is utilitarian only when it leads to measurable performance improvement for the specific model. For a passage to be utilitarian, several conditions are implicitly required: (1) \textit{Knowledge gap}: the model does not already possess sufficient knowledge to answer correctly without the passage; (2) \textit{Relevance}: the passage contains information relevant to answering the query; (3) \textit{Comprehensibility}: the model can effectively interpret and leverage the passage during generation. If any condition fails, the passage would not yield a performance gain.

\noindent\subsection{Discussion on Definition Limitations} We acknowledge several limitations of our operational definition while defending its appropriateness as an initial exploration:

(1) \textit{Single-passage vs. passage combinations}: Our definition focuses on pointwise utility, measuring the marginal contribution of each passage independently. In real-world RAG, especially multi-hop scenarios, passages may exhibit complementarity or redundancy, where the utility of one passage depends on the presence of others. We acknowledge this limitation and position our work as an initial step toward understanding LLM-specific utility. Studying single-passage utility first provides a controlled setting to establish whether utility itself is model-dependent; this foundation is essential before investigating more complex passage interactions, which we leave for future work.

(2) \textit{Performance utility vs. broader utility dimensions}: Our definition focuses primarily on whether a passage improves answer correctness. We acknowledge that in broader RAG contexts, retrieved evidence serves additional purposes beyond correctness, such as providing factual grounding for attribution, enhancing response robustness, and supporting interpretability for trust and transparency. Even when an LLM can answer correctly, retrieved passages may still offer non-performance utility in these dimensions. Our current work centers on performance utility as it directly impacts RAG's core objective of generating accurate answers; broader utility dimensions represent valuable future directions.

Following the definition above, we construct gold utilitarian passages for specific LLM using a binary utility formulation. We adopt this approach for three reasons: it provides clear and unambiguous criteria for annotation, is feasible to implement systematically at scale, and captures the fundamental distinction of whether a passage helps the LLM improve its answer, serving as a foundation for future studies on graded or probabilistic utility.

\section{Investigating LLM-Specific Utilitarian Passages versus Human Evidence} 
\label{sec:human_llm}
\begin{table}[t]
  \centering
    \renewcommand{\arraystretch}{0.8}
       \setlength\tabcolsep{2.0pt}
  \caption{The average number of annotated gold utilitarian passages for various LLMs across three datasets.}
    \begin{tabular}{lccccccccc}
    \toprule
    \multicolumn{1}{c}{\multirow{3}[4]{*}{LLM}} & \multicolumn{3}{c}{NQ} & \multicolumn{3}{c}{TQ} &  \multicolumn{3}{c}{MQ}  \\
\cmidrule(r){2-4} \cmidrule(r){5-7}  \cmidrule(r){8-10}   & $RP$ & $HGP$ & $MCP$ & $RP$& $HGP$  & $MCP$& $RP$& $HGP$ &  $MCP$ \\
    \midrule
    Llama3.1-8B & 1.70  & 0.67  & 1.92  & 1.37  & 0.33  & 1.55  & 1.78  & 0.45  & 1.88  \\
 Qwen3-8B & 2.00  & 0.82  & 2.28  & 2.30  & 0.56  & 2.58  & 1.97  & 0.51  & 2.09  \\

    Qwen3-14B &1.87  & 0.78  & 2.13  & 1.77  & 0.46  & 2.01  & 1.87  & 0.53  & 2.00  \\
  Qwen3-32B & 1.85  & 0.73  & 2.08  & 1.75  & 0.40  & 1.95  & 2.05  & 0.52  & 2.17  \\
    \bottomrule
    \end{tabular}%
  \label{tab:annotation_number}%
\end{table}%

\begin{figure}[t]
    \centering
     \includegraphics[width=\linewidth]{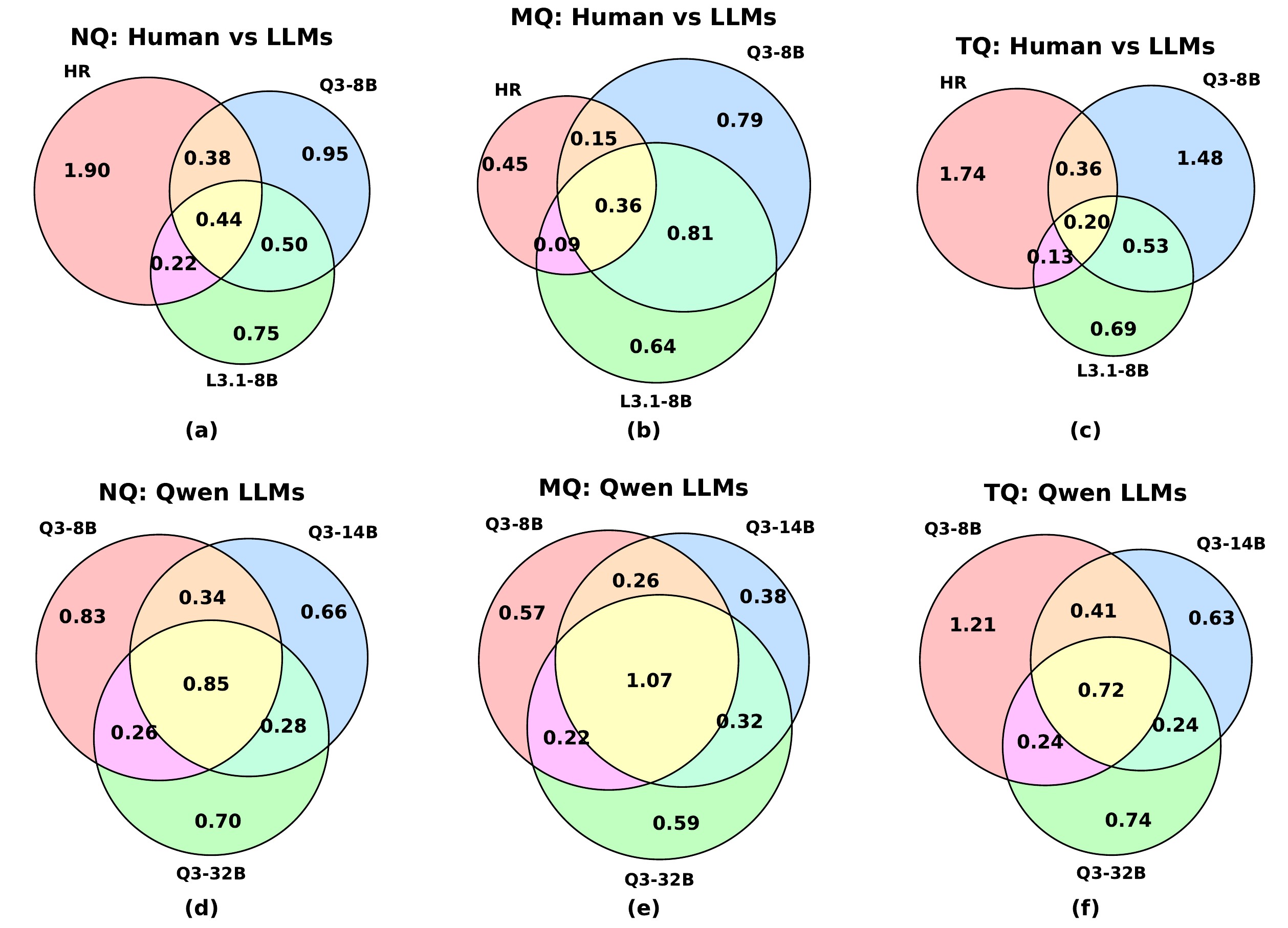}
    \caption{(a), (b), and (c): The average number of overlapping passages between gold utilitarian passages of specific LLMs (candidates: both retrieval results and human-annotated gold passages) and human-annotated gold passages. (d), (e), and (f): The average number of overlapping gold utilitarian passages between different Qwen3 models.}
    \label{fig:human_llm_overlap}
\end{figure}

\begin{table*}[t]
  \centering
  \caption{RAG accuracy (\%) using different evidence types across datasets and LLMs. Results are grouped by dataset (NQ, TriviaQA, MS MARCO-FQA) and LLM. Bold values indicate the best performance for each LLM on the same dataset. Superscript $^{\dagger}$ indicates that the corresponding RAG method yields a statistically significant improvement over the human-annotated gold passages RAG baseline, according to the paired t-test ($p < 0.05$).}
  \renewcommand{\arraystretch}{0.8}
  \setlength\tabcolsep{2pt}
    \begin{tabular}{lcccccccccccc}
    \toprule
    \multirow{2}{*}{Evidence} 
             & \multicolumn{4}{c}{NQ} 
             & \multicolumn{4}{c}{TriviaQA} 
             & \multicolumn{4}{c}{MS MARCO-FQA} \\
    \cmidrule(r){2-5} \cmidrule(r){6-9} \cmidrule(r){10-13}
             & L3.1-8B & Q3-8B & Q3-14B & Q3-32B
             & L3.1-8B & Q3-8B & Q3-14B & Q3-32B
             & L3.1-8B & Q3-8B & Q3-14B & Q3-32B \\
    \midrule
    Without retrieval  & 36.24 & 24.71 & 30.38 & 34.54
              & 75.52 & 59.19 & 68.18 & 71.26
              & 19.13 & 17.47 & 20.69 & 20.82 \\
    $RP$       & 58.12 & 58.41 & 60.73 & 64.08
              & 88.30 & 87.61 & 89.83 & 90.41
              & 40.45 & 42.76 & 45.14 & 47.64 \\ 
     $MCP$    & 61.16 & 59.89 & 63.62 & 67.36
              & 89.74 & 89.12 & 91.23 & 91.64
              & 41.36 & 43.83 & 46.51 & 49.27 \\
    $HGP$     & 70.46 & 67.32 & 71.55 & 73.60
              & 90.67 & 88.04 & 91.25 & 91.42
              & 60.33 & 64.86 & 70.37 & 70.65 \\
    Gold Utilitarian Passages  ($RP$) & 76.17$^\dagger$ & 74.59$^\dagger$ & 76.84$^\dagger$ & 78.39$^\dagger$
              & 95.84$^\dagger$ & 94.72$^\dagger$ & 95.80$^\dagger$ & 96.21$^\dagger$
              & 62.96 & 66.71 & 70.58 & 70.02 \\
    Gold Utilitarian Passages  ($HGP$) & 76.98$^\dagger$ & 74.02$^\dagger$ & 76.45$^\dagger$ & 79.17$^\dagger$
              & 93.92$^\dagger$ & 90.97$^\dagger$ & 93.58$^\dagger$ & 93.95$^\dagger$
              & 63.05$^\dagger$ & 67.40$^\dagger$ & 71.77$^\dagger$ & 71.24$^\dagger$ \\
    Gold Utilitarian Passages ($MCP$) & \textbf{80.30}$^\dagger$ & \textbf{78.60}$^\dagger$ & \textbf{80.68}$^\dagger$ & \textbf{82.16}$^\dagger$
                 & \textbf{97.01}$^\dagger$ & \textbf{95.84}$^\dagger$ & \textbf{97.20}$^\dagger$ & \textbf{97.18}$^\dagger$
                 & \textbf{67.61}$^\dagger$ & \textbf{71.21}$^\dagger$ & \textbf{76.24}$^\dagger$ & \textbf{75.46}$^\dagger$ \\
    \bottomrule
    \end{tabular}
  \label{tab:rq1_rag}
\end{table*}

A user’s satisfaction with search results depends on their personal knowledge, meaning the same passage can be useful to one person but not another. Similarly, in RAG, the utility of a retrieved passage for answering a query can vary across different LLMs due to differences in model knowledge and comprehension. 
Despite this, most RAG and retrieval systems are benchmarked against human-annotated, static datasets, implicitly assuming that human gold passages are also optimal for LLMs.  This prevailing practice raises a critical question: Are human-annotated gold passages truly optimal as evidence for LLMs in RAG? Or do LLMs require their own tailored, model-specific utilitarian passages to achieve the best performance? 
To investigate this, we systematically examine how human-annotated relevant passages relate to ground-truth utilitarian passages for different LLMs. 

\subsection{Experiment Setup} 
\heading{Source Datasets}
In this work, we focus on the factual questions, as non-factual queries are very difficult to evaluate \cite{DBLP:conf/sigir/BolotovaBSCS22}.  We employ several widely used factoid question answering datasets for our analysis. 
Specifically, we utilize two datasets from the KILT benchmark \cite{petroni2020kilt}: Natural Questions (NQ), and TriviaQA. Additionally, we incorporate the MS MARCO QA development set. All queries in these datasets are factoid questions, where the answer is typically an entity. 
\begin{itemize}[leftmargin=*,itemsep=0pt,topsep=0pt,parsep=0pt]
    \item \textbf{Natural Questions (NQ) \cite{kwiatkowski2019natural}}: This dataset comprises real user queries from Google Search. To enhance provenance coverage, KILT conducted an Amazon Mechanical Turk campaign for the development and test sets, increasing the average number of provenance pages per question from 1 to 1.57. 
    \item \textbf{TriviaQA (TQ) \cite{joshi2017triviaqa}}: A collection of question-answer-evidence triples where evidence documents are automatically collected from Wikipedia. KILT uses only the Wikipedia-based portion of this dataset. 
    \item \textbf{MS MARCO-FQA (MQ) \cite{nguyen2016ms}}: The development set comprises queries sampled from Bing's search logs, each accompanied by human-annotated passages. Queries are categorized by type: {LOCATION, NUMERIC, PERSON, DESCRIPTION, ENTITY}. We focused on the factual questions within the MS MARCO dataset (approximately half of all queries). 
    Unlike the NQ dataset,  MS MARCO's ground-truth answers are sentences rather than specific entities.  
    To address this, we employed the Qwen3-32B model to extract precise entity answers from the filtered MS MARCO data. This newly processed collection is termed the MS MARCO-FQA dataset.
\end{itemize}

\heading{Retriever} 
For the datasets from the KILT benchmark as well as 2WikiQA, the corpus is the Wikipedia dumps and then split into DPR \cite{karpukhin2020dense} 100-word format as passages, resulting in about 34M passages, which is provided by DPR\footnote{\url{https://dl.fbaipublicfiles.com/ur/wikipedia_split/psgs_w100.tsv.gz}}. 
For the MS MARCO-FQA, we directly use the MS MARCO v1 passage corpus \cite{nguyen2016ms}, which contains about 8.8M passages. 
We utilize the well-performing BGE-M3 \cite{multi2024m3} as our retriever and retrieve top-20 results for all six datasets. 

\heading{Analyzed LLMs}
We analyze four widely used LLMs from two families:  Llama3.1-8B-Instruct (denoted as L3.1-8B) \cite{dubey2024llama}, Qwen3-8B (denoted as Q3-8B), Qwen3-14B (denoted as Q3-14B), and Qwen3-32B \cite{qwen3technicalreport} (denoted as Q3-32B). 
Qwen spans 8B–32B for studying scale effects, while Llama3.1-8B provides a cross-family baseline.  By default, Qwen3's think function is disabled. 
To ensure reproducibility, the temperature for all LLMs in this study was set to 0. 
We leave the investigation on more LLM families and under non-zero temperatures with multiple sampling in future work.

\heading{Candidate Pool Option}
To facilitate a comprehensive evaluation of gold utilitarian passage annotation, we employ three distinct candidate pools for constructing gold utilitarian passages of four LLMs: Llama3.1-8B, Qwen3-8B, Qwen3-16B, and Qwen3-32B:  
(1) \textbf{Top-20 Retrieved Passages ($RP$)}: This pool reflects a typical scenario in real-world RAG systems, where only retrieval results are available, and human annotations are not always accessible.  
(2) \textbf{Human-Annotated Gold Passages ($HGP$)}: Utilizing human-annotated gold passages allows for direct comparison with human judgments and assessment of alignment with human annotation quality.  
(3) \textbf{Merged Candidate Pool ($MCP$)}: By merging the top-20 retrieved passages with the human-annotated gold passages, we create an expanded candidate pool, offering a more complete annotation set for gold utilitarian passages. 
In this section, we focus on analyzing the RAG performance with respect to the relationship between human-annotated gold passages and LLM-specific utility.  
Our utility definition focuses on the pointwise utility, the utility of a passage to a given LLM considered independently of other passages.  
To ensure a fair comparison with human-annotated gold passages, we restrict the experiments in this section to three single-hop datasets: NQ, TriviaQA, and MS MARCO-FQA.

\heading{Known and Unknown Query Definition}
Based on our utility definition, if an LLM can already answer correctly without passages, additional evidence provides limited gain. We thus categorize queries into \textbf{known queries} (answerable without passages) and \textbf{unknown queries} (requiring external information).

\subsection{Statistics of Gold Utilitarian Passages} 
We present the statistics of the number of gold utilitarian passages and the overlap between gold utilitarian passages and human-annotated gold passages. 

\heading{$GUP$ Number Statistics}
Table \ref{tab:annotation_number} shows the annotated numbers of gold utilitarian passages for different LLMs. We find that incorporating human-annotated gold passages into the automatically annotated set resulted in only a marginal increase in the number of golden passages, suggesting that human-annotated gold passages are not always labeled as gold utilitarian passages for specific LLMs.

\begin{figure}[t]
    \centering
     \includegraphics[width=\linewidth]{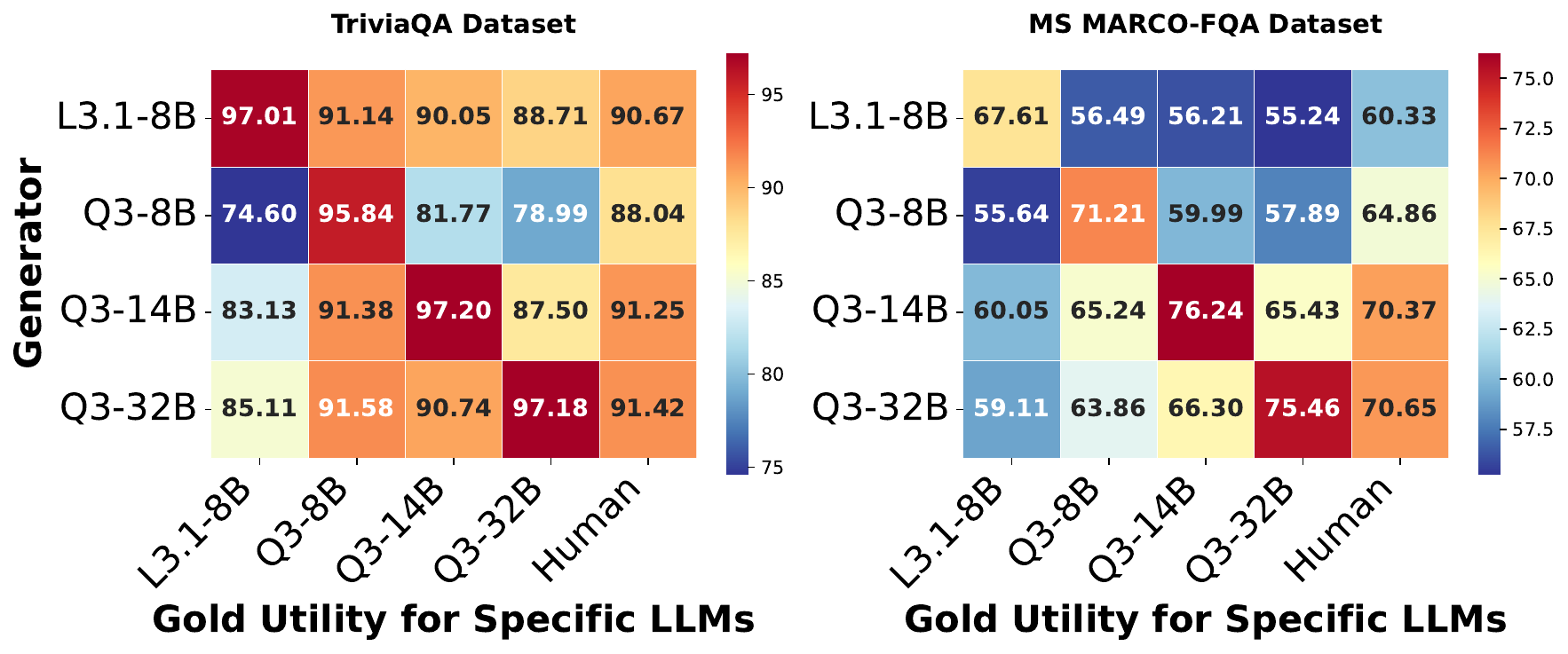}
    \caption{RAG accuracy (\%) of LLMs with gold utilitarian passages (where the candidate set is $MCP$) from different LLMs. The results on the NQ dataset are shown in Figure \ref{fig:utility_relevance} (b).}
    \label{fig:rq1_heatmap_union}
\end{figure}

\begin{figure*}[t]
    \centering
     \includegraphics[width=\linewidth]{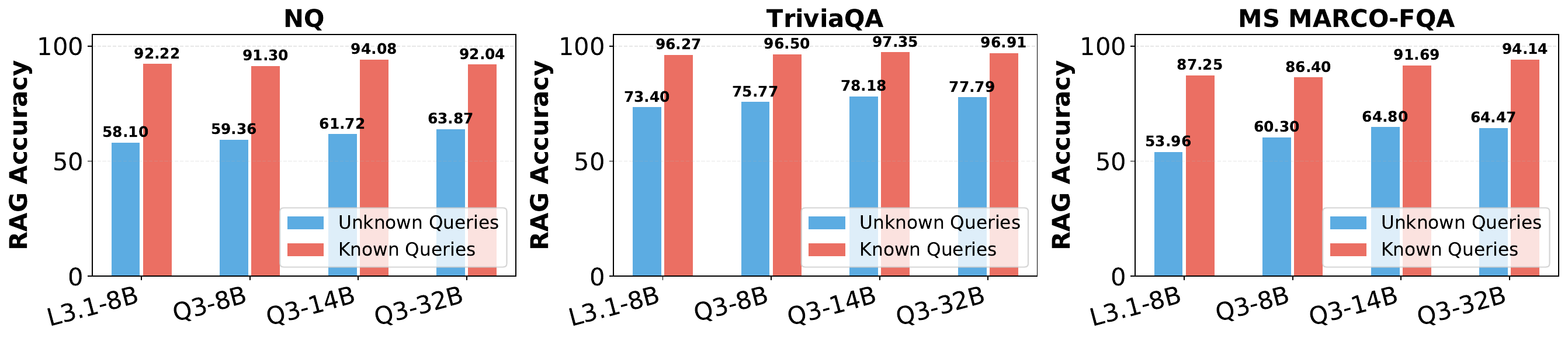}
    \caption{RAG accuracy (\%) of LLMs using human-annotated gold passages across different query types. Known queries are those that the LLM can answer correctly without retrieval (RAG performance is approximately 100\% without retrieval), while unknown queries are those for which the LLM cannot provide a correct answer without retrieval (RAG performance is approximately 0\% without retrieval).}
    \label{fig:human_known_unknown}
\end{figure*}

\begin{figure}[t]
    \centering
     \includegraphics[width=\linewidth]{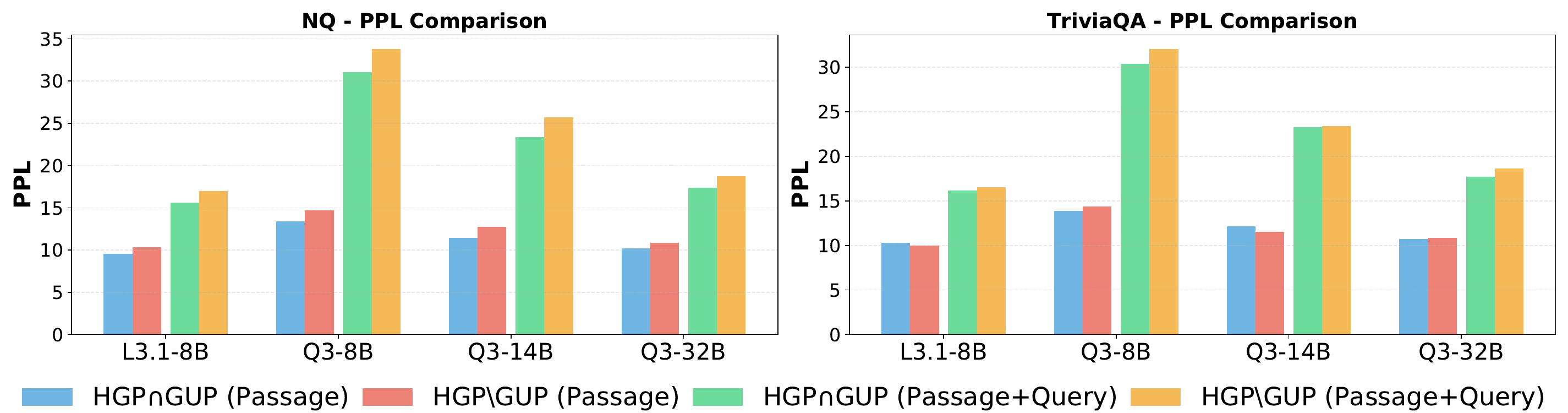}
    \caption{Perplexity (PPL) of LLMs on human-annotated gold passages that have been labeled as gold utility (i.e., $HGP \cap GUP$) and those not labeled as gold utility (i.e., $HGP \setminus GUP$), respectively. }
    \label{fig:human_ppl}
\end{figure}

\heading{Overlap Between $GUP$ and $HGP$}
To investigate the difference between human-annotated gold passages and the LLM-specific gold utilitarian passages, we compute the intersection between these two sets, as illustrated in Figure \ref{fig:human_llm_overlap}.
Our observations are as follows: 
(1) As shown in Figure \ref{fig:human_llm_overlap} (a–c), on average, only about half of the human-annotated gold passages are included in the LLM-specific gold utilitarian passages for both the NQ and MS MARCO-FQA datasets. Regarding the gold utilitarian passages not annotated by humans, manual inspection reveals that most of these passages are genuinely useful for answering the questions. This may be due to their absence from the annotation pool during the human annotation process.
(2) As illustrated in Figure \ref{fig:human_llm_overlap} (d–f), LLMs from the same family exhibit a higher degree of overlap in their utilitarian passages compared to those from different families.

\subsection{Experimental Results}


\heading{Human-Annotated Gold Passages vs. LLM-Specific Gold Utilitarian Passages}
Table \ref{tab:rq1_rag} compares the RAG using these different references. We can observe that: 
(1) The highest RAG performance across all evaluated LLMs and datasets is achieved using gold passages of utility constructed from a union of candidate passages. This result underscores the critical importance of tailoring utility assessment to the specific LLM used for answer generation.  
(2) Using these ground-truth utilitarian passages—derived from top-20 retrieval results—yields a significantly greater improvement in answer performance compared to using all top-20 results, as well as the union candidate. 
Furthermore, this LLM-specific utility generally surpasses the performance of human-annotated general utility. This pattern suggests that the quality of retrieval results, as validated by the ground-truth utilitarian passages, is the dominant factor for a given dataset. Collectively, these findings establish the performance upper bound for the utility judgments on the candidate passages. 
(3) While using the $MCP$ candidate leads to better answer performance than the standard top-20 results, it remains worse than using human-annotated gold passages. This performance gap is particularly pronounced on the NQ and MS MARCO-FQA datasets, indicating that the presence of irrelevant or ``noisy'' passages can substantially impair LLM performance during answer generation. 

\heading{Transferability of Gold Utilitarian Passages Across LLMs}
\label{sec:rq1:trans}
Human-annotated general utility assumes that the annotated passages are universally useful for all LLMs. This naturally raises the question of whether gold utilitarian passages selected for a specific LLM are also effective for other LLMs. To explore this, we conduct a transfer experiment with gold utilitarian passages: passages constructed for one LLM are used by other LLMs to answer questions. 
We perform transfer experiments, as shown in  Figure \ref{fig:rq1_heatmap_union} (with additional results on the NQ dataset presented in Figure \ref{fig:utility_relevance} (b)). 
We can observe that: 
(1) The LLM-specific gold utilitarian passages are not shareable with other LLMs.
This finding underscores the necessity of LLM-specific calibration and challenges the practice of employing a static set of annotated passages to evaluate retrieval quality across diverse LLMs. 
(2) LLMs from the same family (e.g., the Qwen3 series) have more aligned information needs. This is shown by the fact that Qwen3 models perform worse when using gold passages constructed for Llama3.1-8B than when using passages constructed for other Qwen models. 
(3) Figure \ref{fig:rq1_heatmap_union} shows a consistent performance hierarchy across different LLMs: an LLM achieves its best results with its own utilitarian passages, followed by general human-annotated gold passages, and performs worst with the utilitarian passages optimized for a different LLM. This pattern confirms that while human-annotated relevant passages may not be optimal for any single LLM, they capture a general utility that is robust, widely applicable, and effective.   

\begin{table}[t]
  \centering
  \small
  \caption{LLM-specific utility judgments performance (F1, \%) of different utility-based selection methods. Bold indicates the best performance for the specific LLM. ``PA'' stands for ``Pseudo answer''.  ``2QA'', and ``HQA'' refer to the   2WikiQA and HotpotQA datasets, respectively. }
  \renewcommand{\arraystretch}{0.8}
   \setlength\tabcolsep{2pt}
    \begin{tabular}{llllllcll}
    \toprule
         Method     & NQ & HQA & 2QA & TQ & MQ &  FEVER & Overall \\
    \midrule
    \multicolumn{8}{c}{Llama3.1-8B} \\
    \midrule
   PVS  ($w/o$ $PA$)  & 40.28  & 40.28  & 36.93  & 58.07  & 37.43  & 50.25   & 45.90 \\
    PVS ($with$ $PA$)   & 48.61  & 48.61  & 39.70  & \textbf{63.05} & 44.29  & 42.45 & 47.44  \\
    \midrule
    LVS ($w/o$ $PA$)  & 47.86  & 47.86  & 38.59  & 58.58  & 44.12  & 52.13   & 48.40  \\
    LVS ($with$ $PA$)   & \textbf{49.88 } & \textbf{49.88 } & \textbf{39.91 } & 60.28  & \textbf{47.13} & \textbf{52.72} & \textbf{49.84}    \\
    \midrule
     \multicolumn{8}{c}{Qwen3-8B} \\
     \midrule
    PVS ($w/o$ $PA$)  & 48.90  & 48.90  & 25.88  & 63.95  & 45.29  & 45.05  & 44.13    \\
    PVS ($with$ $PA$)   & 56.66  & 56.66  & 26.51  & 69.40  & 49.14  & 45.83 & 46.06  \\
    \midrule
   LVS ($w/o$ $PA$)  & 56.05  & 56.05  & 37.78  & 67.83  & 49.50  & \textbf{59.24}  & 53.23  \\
    LVS ($with$ $PA$)   & \textbf{58.37} & \textbf{58.37} & \textbf{37.95} & \textbf{69.66} & \textbf{52.24} & 59.22  & \textbf{53.99}  \\
    \midrule
    \multicolumn{8}{c}{Qwen3-14B} \\
    \midrule
    PVS ($w/o$ $PA$)  & 49.66  & 49.66  & 17.10  & 63.08  & 47.02  & 56.83  & 41.32   \\
     PVS ($with$ $PA$)   & \textbf{57.86} & \textbf{57.86} & 23.56  & 71.59  & 48.91  & 62.24 & 47.02     \\
     \midrule
    LVS ($w/o$ $PA$)  & 55.14  & 55.14  & 39.91  & 67.68  & 47.50  & 63.05   & 52.88   \\
    LVS ($with$ $PA$)   & 57.27  & 57.27  & \textbf{40.67} & \textbf{69.70} & \textbf{50.15} & \textbf{63.64}  & \textbf{54.13}  \\
    \midrule
    \multicolumn{8}{c}{Qwen3-32B} \\
    \midrule
     PVS ($w/o$ $PA$)  & 50.17  & 50.17  & 17.62  & 62.29  & 47.92  & 40.33 & 38.81   \\
     PVS ($with$ $PA$)   & 54.79  & 54.79  & 24.28  & \textbf{67.64 } & 48.65  & \textbf{49.34}  & 43.85  \\
     \midrule
  LVS ($w/o$ $PA$)  & 56.61  & 56.61  & \textbf{39.89 } & 64.24  & 51.29  & 44.20  & \textbf{50.83}    \\
     LVS ($with$ $PA$)   & \textbf{55.87} & \textbf{55.87} & 39.88  & 64.62  & \textbf{53.36} & 44.15 & 50.79    \\
    \bottomrule
    \end{tabular}%
  \label{tab:rq2_set}%
\end{table}%

\heading{Human-Annotated Passage Effectiveness on Known and Unknown Queries}
We further analyze the impact of human-annotated gold passages on these two query types, as shown in Figure \ref{fig:human_known_unknown}. Our observations are as follows: 
(2) Unknown queries: For queries that LLMs cannot answer correctly without retrieval, providing human-annotated gold passages yields consistently lower performance compared to known queries. Stronger LLMs, such as Qwen3 14B and Qwen3 32B, demonstrate better performance on their respective unknown queries when given human-annotated gold passages, compared to weaker models like Llama3.1-8B and Qwen3-8B. These results suggest that passages easily understood by humans may not be equally interpretable by LLMs, although stronger LLMs tend to benefit more from such passages. This aligns with findings in the query suggestion task in information retrieval \cite{block2022counterfactual}, where human-suggested queries do not always lead to higher retrieval performance. 
(2) Known queries: For queries that LLMs can answer correctly without retrieval, even human-annotated gold passages can sometimes lead to slight performance degradation.


\heading{Perplexity Highlights LLM-Specific Comprehension Gaps in Utilizing $HGP$}
From the Figure \ref{fig:human_llm_overlap}, we can find that not all human-annotated gold passages are labeled as gold utilitarian passages for specific LLMs. 
As for why some human-annotated gold passages are absent from an LLM’s gold utilitarian set, we hypothesize that the discrepancy stems from the understanding capabilities of LLMs. 
Figure \ref{fig:human_ppl} displays the perplexity of various LLMs on different kinds of datasets. 
Our findings can be summarized as follows: 
(1) Across human-annotated gold passages, those included in an LLM’s gold utilitarian set exhibit lower perplexity than those excluded, with the strongest effect observed on NQ. Likewise, the joint perplexity of a query paired with its gold utilitarian passage is consistently lower than that of the same query paired with a non-utility passage, a pattern that holds across both datasets. 
Even among semantically similar human-annotated gold passages containing comparable evidence, LLMs assign different perplexities and derive different utility, indicating that utility depends not only on content but also on the model’s ease of processing it. 
These results suggest that perplexity captures aspects of readability and comprehension: passages that a model finds easier to parse are more likely to yield utility, which helps explain why some human-annotated gold passages are absent from LLM-specific utility sets. This observation aligns with prior findings on source bias \cite{wang2025perplexity}, where, under comparable semantics, retrieval models preferentially give lower-perplexity documents a higher relevance score, reflecting better model understanding. 
(2) Within a given model family, stronger LLMs, those achieving better RAG performance like Qwen3-32B, consistently exhibit lower perplexity on both the passages and the query-passage pairs.

\begin{table*}[t]
  \centering
    \renewcommand{\arraystretch}{0.8}
  \caption{Ranking performance (NDCG@5, \%) of various LLM-specific utility-based ranking methods.  Bold values indicate the best performance for each LLM. Superscripts $^{\dagger}$ and $^{\ddagger}$ denote statistically significant improvements over all other methods and over the likelihood method within the same block using the paired t-test ($p < 0.05$), respectively.}
    \begin{tabular}{lllllcll}
    \toprule
     Method & NQ & HotpotQA & 2WikiQA & TriviaQA & MS MARCO-FQA & FEVER & Overall\\
    \midrule
    \multicolumn{8}{c}{Llama3.1-8B-Instruct} \\
    \midrule
    Initial Retrieval & 52.72 & 51.11 & 43.64 & 56.68 & 53.35 & 66.39 & 52.29  \\
AttR & 29.03 & 27.95 & 38.79 & 33.69 & 28.23 & 60.12 & 38.24 \\
 Likelihood & \textbf{62.66}$^\dagger$ & \textbf{64.67}  & 51.73 & \textbf{78.25}$\dagger$   & 54.01 & \textbf{71.75}$^\dagger$ & \textbf{61.28}$^\dagger$  \\
 LVR ($w/o$ Pseudo answer) & 57.76 & 61.09 & 51.67 &68.32 & 53.72 & 69.87 & 58.89  \\
 LVR ($w/$ Pseudo answer) & 60.36 & 63.67& \textbf{53.54}$^{\dagger\ddagger}$  & 69.73 & \textbf{55.06}  & 69.59 & 60.44  \\
    \midrule
    \multicolumn{8}{c}{Qwen3-8B} \\
    \midrule
    Initial Retrieval & 54.47 & 49.46 & 38.06 & 59.89 & 55.32 & 71.05 & 52.11 \\
 AttR & 45.69 & 43.22 & 34.07 & 48.98 & 42.37 & 62.78 & 44.50 \\
Likelihood & 64.77 & 66.42 & 48.94 & \textbf{81.27}$^\dagger$  & 52.44 & \textbf{82.95}$^\dagger$ & 63.80  \\
LVR ($w/o$ Pseudo answer) & 68.25$^\ddagger$ & 66.70& 48.31 & 79.52 & 59.61$^\ddagger$ & 77.82 & 63.77  \\
LVR ($w/$ Pseudo answer) & \textbf{69.83}$^{\dagger\ddagger}$  & \textbf{68.08}${^{\dagger\ddagger}}$  & \textbf{49.53}  & 79.53 & \textbf{60.02}$^\ddagger$  & 77.78 & \textbf{64.57}$^{\dagger\ddagger}$  \\
    \midrule
    \multicolumn{8}{c}{Qwen3-14B} \\
    \midrule
    Initial Retrieval & 54.04 & 49.19 & 37.03 & 59.37 & 53.89 & 70.23 & 50.65  \\
 Likelihood & 64.41 & 67.51 & \textbf{55.91}$^\dagger$  & \textbf{84.26}$^\dagger$  & 51.98 & 74.18 & 64.27  \\
 LVR ($w/o$ Pseudo answer) & 68.47$^\ddagger$& 67.83 & 50.92 & 80.92& \textbf{60.35}$^\ddagger$  & 79.32$^\ddagger$ & 64.49  \\
LVR ($w/$ Pseudo answer) & \textbf{69.51}$^{\dagger\ddagger}$ & \textbf{68.59}$^\ddagger$  & 52.22 & 80.95 & 60.34$^\ddagger$ & \textbf{79.57}$^\ddagger$  & \textbf{65.21}$^{\dagger\ddagger}$  \\
    \midrule
    \multicolumn{8}{c}{Qwen3-32B} \\
    \midrule
   Initial Retrieval & 53.93 & 48.50  & 40.06 & 59.89 & 54.15 & 69.96 & 51.19  \\
  Likelihood & 61.01 & 66.58 & \textbf{55.89}$^\dagger$  & \textbf{81.02}$^\dagger$  & 50.84 & \textbf{73.40}  & 62.71  \\
 LVR ($w/o$ Pseudo answer) & 66.87$^\ddagger$ & 67.40 & 52.81 & 79.49 & \textbf{59.29}$^\ddagger$ & 72.92 & 63.22  \\
  LVR ($w/$ Pseudo answer) & \textbf{67.72}$^{\dagger\ddagger}$ & \textbf{68.95}$^{\dagger\ddagger}$ & 53.96 & 79.50  & 59.15$^\ddagger$ & 73.30  & \textbf{64.00}$^{\dagger\ddagger}$ \\
  \bottomrule
    \end{tabular}%
  \label{tab:rq2_ranking}%
\end{table*}%

\section{LLM-Specific Utility Judgment}
Given that utilitarian passages vary across LLMs, a key research question is how to help each LLM identify its useful evidence efficiently. We propose this task to facilitate such research: our utility labels serve as ground truth for evaluating or training methods that can generalize to new LLMs. 

\subsection{Task Description} 
The \textbf{LLM-specific utility judgment task} aims to select supporting evidence that is specifically utilitarian for a given large language model (LLM). 
Formally, given a question $q$ and a set of $N$ retrieved candidate passages $\mathcal{D} = \{d_1, d_2, \ldots, d_N\}$ from the corpus $\mathcal{C}$, the goal is to select a subset $\mathcal{D}_u \subseteq \mathcal{D}$ that is utilitarian for the target LLM $L$ in answering $q$. 
Given the gold utilitarian passages for each LLM, defined in Equation \eqref{eq:utility-def} \& \eqref{eq:utilitarian-set}, we evaluate the ability of LLMs to assess the utility of retrieved passages using two approaches: set-based evaluation, i.e., $\mathcal{D}_u$ is a set of utilitarian passages selected from $\mathcal{D}$, and ranking-based evaluation, i.e., $\mathcal{D}_u$ is a ranked list based on utility. 


\begin{table}[t]
  \centering
  \caption{Dataset statistics.}
  \renewcommand{\arraystretch}{0.8}
   \setlength\tabcolsep{2.5pt}
    \begin{tabular}{lllllll}
    \toprule
          & NQ    & HotpotQA & 2WikiQA & TQ & MQ & FEVER \\
    \midrule
    \#Queries &  2837 &5600 & 12576 & 5359 & 3199 &10245 \\
    \#Corpus & 32M   & 32M   & 32M   & 32M   & 8.8M  & 32M \\
    \bottomrule
    \end{tabular}%
  \label{tab:data_sta}%
\end{table}%
\subsection{Benchmark Construction} 
We restricted evaluation to three single-hop datasets to ensure fair comparison with human annotations in Section \ref{sec:human_llm}. 
We focus on pointwise utility, and our gold annotations identify passages that are individually useful to each target LLM. While this formulation may not capture utility that emerges from combining multiple passages, some multi-hop questions can still be answered from a single passage when it contains sufficient evidence. Such instances therefore remain valid for evaluating pointwise utility, making the inclusion of multi-hop data consistent with our task definition. 
Thus we use more comprehensive datasets (NQ, HotpotQA, 2WikiQA, TriviaQA, MS MARCO-FQA, FEVER) covering single- and multi-hop questions in this section to construct the benchmark, i.e., \textbf{\benchmark} (LLM-Specific Utility Benchmark). 

\heading{Additional Source Datasets} In addition to the single-hop datasets described earlier, we include the following multi-hop datasets: 
\begin{itemize}[leftmargin=*,itemsep=0pt,topsep=0pt,parsep=0pt]
    \item \textbf{HotpotQA (HQ) \cite{yang2018hotpotqa}}: This dataset provides question-answer pairs along with human-annotated supporting sentences. We adopt the full wiki setting, which requires systems to retrieve and reason over the entire Wikipedia.
    \item \textbf{FEVER \cite{thorne2018fever}}: A large-scale dataset designed for claim verification, where systems must retrieve sentence-level evidence to determine whether a claim is supported or refuted.
    \item \textbf{2WikiQA (2Wiki) \cite{xanh2020_2wikimultihop}}: A multi-hop question answering dataset that incorporates both structured and unstructured data. 
\end{itemize}
Table \ref{tab:data_sta} shows the detailed dataset statistics for all datasets used in \benchmark.  
Moreover, in real-world applications, human-annotated gold passages are not always available. 
Therefore, we use the top-20 retrieval results from BGE-M3 \cite{multi2024m3} as the annotation pool for constructing gold utilitarian passages for specific LLMs in this section.   


\begin{table*}[t]
  \centering
  \small
  \caption{RAG accuracy (\%) with various evidence sources across multiple datasets and LLMs. \textbf{Bold} numbers indicate the best result within the same block.  Superscripts ${\dagger}$, ${\star}$, and ${\ddagger}$ mark statistically significant improvements over RAG with other methods' results, RAG with likelihood-ranked results, and over RAG with initial retrieval results within the same block using the paired t-test ($p < 0.05$), respectively.}
    \renewcommand{\arraystretch}{0.8}
   \setlength\tabcolsep{1.5pt}
    \begin{tabular}{llllllllllllll}
    \toprule
    \multirow{2}[3]{*}{Source} & \multicolumn{2}{c}{NQ} & \multicolumn{2}{c}{HotpotQA} & \multicolumn{2}{c}{TQ} & \multicolumn{2}{c}{MQ} & \multicolumn{2}{c}{FEVER} & \multicolumn{2}{c}{2WikiQA} & \multicolumn{1}{c}{\multirow{2}[3]{*}{Overall}} \\
\cmidrule(r){2-3}  \cmidrule(r){4-5}   \cmidrule(r){6-7} \cmidrule(r){8-9} \cmidrule(r){10-11} \cmidrule(r){12-13}               & \multicolumn{1}{l}{Unknown} & \multicolumn{1}{l}{Known} & \multicolumn{1}{l}{Unknown} & \multicolumn{1}{l}{Known} & \multicolumn{1}{l}{Unknown} & \multicolumn{1}{l}{Known} & \multicolumn{1}{l}{Unknown} & \multicolumn{1}{l}{Known} & \multicolumn{1}{l}{Unknown} & \multicolumn{1}{l}{Known} & \multicolumn{1}{l}{Unknown} & \multicolumn{1}{l}{Known} &  \\
\midrule
    \multicolumn{13}{c}{Llama3.1-8B-Instruct (Utility-Based Selection)}                                                        &  \\
    \midrule
     Initial Retrieval (Top-20) & 42.62  & 85.41  & 25.92  & 80.82  & 70.05  & 94.22  & 31.19  & 79.58  & 69.26  & 92.61  & 24.97  & \textbf{62.85}  & 58.63  \\
    PVS ($w/$ A) & 43.45  & \textbf{85.89}  & 27.09  & \textbf{82.79}$^\ddagger$  & \textbf{72.26}$^\ddagger$  & \textbf{95.21}$^\ddagger$  & 31.23  & 80.23  & 56.72  & 90.75  & 23.96  & 60.63  & 57.52  \\
    LVS ($w/$ A) & \textbf{44.06}$^\ddagger$  & 85.70  & \textbf{28.62}$^\dagger$  & \textbf{82.79}$^\ddagger$  & 71.11  & 94.86$^\ddagger$  & \textbf{31.43}  & \textbf{81.37}  & \textbf{71.77}$^{\dagger\ddagger}$  & \textbf{93.29}$^{\dagger\ddagger}$  & \textbf{25.19}  & 62.11  & \textbf{59.46}$^{\dagger\ddagger}$ \\
    \midrule
    \multicolumn{13}{c}{Llama3.1-8B-Instruct (Utility-Based Ranking)}                                                          &  \\
    \midrule
    Retrieval (Top-5) & 43.56  & 81.03  & 25.66  & 79.16  & 65.93  & 92.51  & 30.34  & 75.16  & 71.16  & 92.83  & 22.85  & 60.79  & 57.53  \\
    AttR (Top-5) & 32.01  & 74.03  & 17.77  & 67.25  & 51.45  & 87.01  & 23.81  & 69.28  & 61.91  & 83.80  & 18.65  & 54.71  & 50.19  \\
    Likelihood (Top-5) & 42.40  & 83.17  & 26.73$^\ddagger$  & \textbf{80.43}  & \textbf{70.58}$^\ddagger$  & \textbf{94.63}$^\ddagger$  & \textbf{30.73}  & 76.96  & 69.65  & 92.37  & 24.78$^\ddagger$  & 61.29  & 58.40$^\ddagger$  \\
    LVR ($w/$ A) (Top-5) & \textbf{44.11}$^\star$  & \textbf{84.05}$^\ddagger$  & \textbf{27.99}$^{\dagger\ddagger\star}$  & 79.48  & 69.82  & 94.29  & 30.65  & \textbf{78.59}$^\ddagger$  & \textbf{72.72}$^{\star}$  & \textbf{93.71}$^{\dagger\ddagger\star}$  & \textbf{24.91}$^\ddagger$  & \textbf{62.27}  & \textbf{59.14}$^{\dagger\ddagger\star}$  \\
    \midrule
    \multicolumn{13}{c}{Qwen3-8B (Utility-Based Selection)}                                                           &  \\
    \midrule
    Retrieval (Top-20) & 49.30  & \textbf{86.16}  & 28.65  & 85.30  & 77.64  & 94.48  & \textbf{34.73}  & \textbf{80.68}  & 77.32  & \textbf{95.99}  & 23.73  & 70.57  & 59.88  \\
    PVS ($w/$ A) & 46.35  & 85.73  & 26.81  & \textbf{87.93}$^\ddagger$  & 75.95  & \textbf{95.33}$^\ddagger$  & 31.36  & 77.10  & 60.65  & 95.67  & 17.42  & \textbf{78.32}$^{\dagger\ddagger}$  & 57.35  \\
    LVS ($w/$ A) & \textbf{49.81}  & 85.02  & \textbf{30.24}$^\dagger$  & 86.18  & \textbf{78.83}$^\dagger$  & 94.14  & 34.36  & 78.71  & \textbf{78.03}  & 95.95  & \textbf{28.44}$^{\dagger\ddagger\star}$  & 72.79  & \textbf{61.40}$^{\dagger\ddagger}$ \\
    \midrule
    \multicolumn{13}{c}{Qwen3-8B (Utility-Based Ranking)}                                                             &  \\
    \midrule
    Initial Retrieval (Top-5) & 46.86  & 83.45  & 24.97  & 82.59  & 72.52  & 92.88  & 32.88  & 73.35  & 76.95  & 95.45  & 22.70  & 68.67  & 58.05  \\
    AttR (Top-5) & 43.63  & 84.02  & 23.49  & 77.95  & 71.10  & 90.94  & 32.20  & 74.78  & 73.78  & 94.72  & 18.83  & 64.51  & 55.75  \\
    Likelihood (Top-5) & 46.63  & 83.88  & 27.10$^\ddagger$  & 84.08$^\ddagger$  & 75.67$^\ddagger$  & 93.53  & 32.35  & 75.67  & 76.28  & \textbf{96.10}$^\ddagger$  & 22.01  & 70.63$^\ddagger$  & 58.63$^\ddagger$  \\
     LVR ($w/$ A) (Top-5)  & \textbf{49.44}$^{\dagger\ddagger\star}$  & \textbf{84.45}  & \textbf{29.32}$^{\dagger\ddagger\star}$  & \textbf{86.70}$^{\dagger\ddagger\star}$  & \textbf{76.77}$^{\dagger\ddagger\star}$  & \textbf{94.32}$^{\dagger\ddagger\star}$  & \textbf{33.71}$^\star$  & \textbf{76.21}$^\ddagger$  & \textbf{78.37}$^{\dagger\ddagger\star}$  & 95.87$^\ddagger$  & \textbf{27.42}$^{\dagger\ddagger\star}$  & \textbf{72.10}$^{\dagger\ddagger}$  & \textbf{60.81}$^{\dagger\ddagger\star}$  \\
    \bottomrule
    \end{tabular}%
  \label{tab:rq2_set_rag}%
\end{table*}%

\heading{Evaluation Setting}
\label{sec:evaluation}
We evaluated current LLM-specific utility judgment methods using different evaluations on the six datasets: 
\begin{itemize}[leftmargin=*,itemsep=0pt,topsep=0pt,parsep=0pt]
    \item \textbf{Utility Judgments Evaluation}. We employ two evaluations using the golden utilitarian passages (queries with empty golden utilitarian passages are excluded): (1) Set-based evaluation: Measured by the F1 score on the selected passage set; (2) Ranking-based evaluation: Measured by Normalized Discounted Cumulative Gain (NDCG) \cite{jarvelin2002cumulated}.  
    For LLM-specific utility judgments, we only consider queries with non-empty gold utilitarian labels. 
    
    \item \textbf{RAG Evaluation}. RAG performance refers to answer generation accuracy ($Acc$), as in prior work \cite{kwiatkowski2019natural, ni2025towards}. 
    For the ranking-based utility judgment method, we use the top-$5$ utility-based ranked passages as evidence for answering, following \citet{zhang2025distilling}.  
\end{itemize} 
As F1 is the harmonic mean of average precision and average recall across all queries (not a query-level F1), it is not suitable for significance testing. 
Hence, significance testing is only conducted on NDCG and accuracy scores using the paired t-test. 



\vspace{-3mm}
\subsection{LLM-Specific Utility Judgment Methods}
\label{sec:self-utility-judgment}
The approaches are categorized according to different evaluation goals, namely \textit{utility-based selection} \cite{zhang2024iterative, zhang2025distilling, zhang2024large} and \textit{utility-based ranking} \cite{shi2023replug, izacard2023atlas, salemi2024evaluating, salemi2025learning, zamani2024stochastic}.  Furthermore, empirical studies \cite{zhang2024large, zhang2024iterative} indicate that utility judgments yield higher performance when judging utility referring to the pseudo-answers generated from retrieved documents.  For all methods, pseudo-answers are generated from the top-$k$ ($k$=20) retrieval results to ensure a fair comparison.

\begin{figure*}
    \centering
    \includegraphics[width=\linewidth]{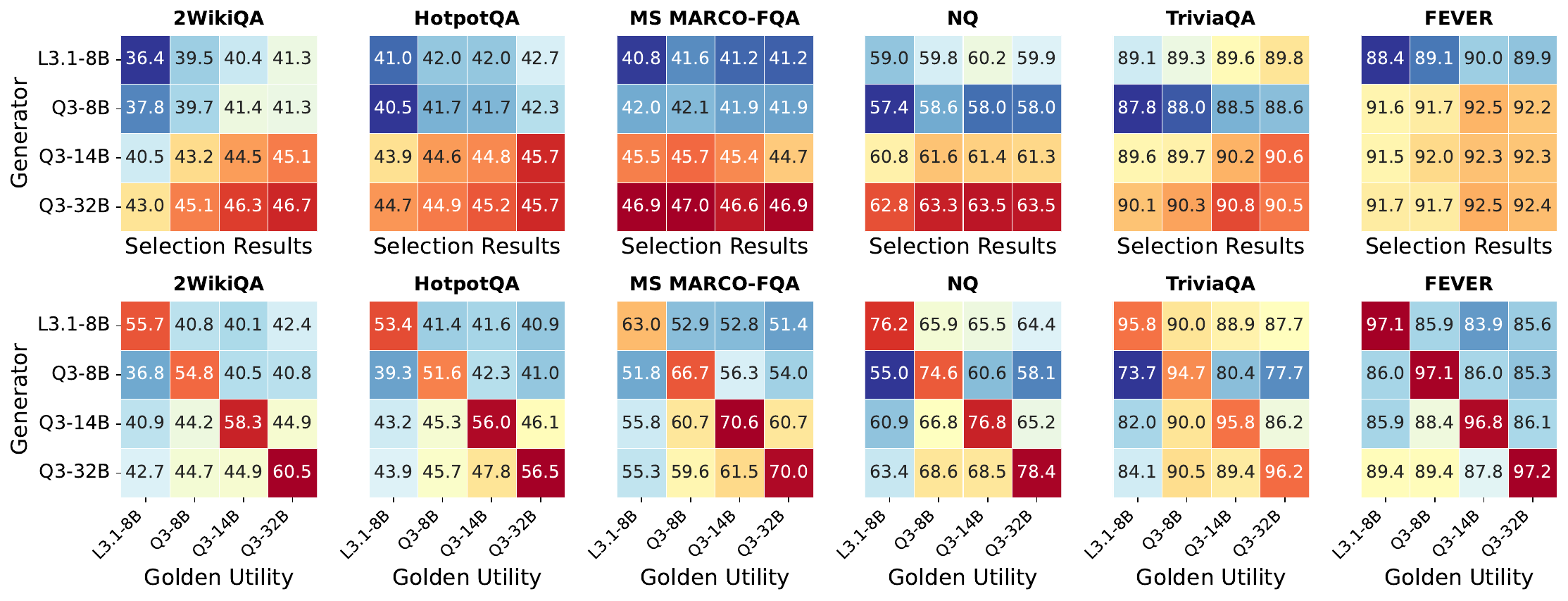}
    \caption{RAG accuracy (\%) with selection results using LVS $with$ Answer (1st row) vs Gold Utilitarian Passage (2nd row).}
    \label{app:fig:top20_trasfer}
\end{figure*}
\textbf{Utility-Based Selection Methods}.
These methods select passages judged to be useful for the LLM, based on a strict utility definition:  
\textsl{``Determine if the passage has utility based on two criteria: 1. Usefulness: The passage must be relevant and contribute to generating a correct, reasonable, and complete answer. 2. Novelty: The useful information should be new, not part of your pre-existing knowledge.''}

Selection methods are categorized by input format and whether a pseudo-answer is provided \cite{zhang2024large}:

\begin{itemize}[leftmargin=*,itemsep=0pt,topsep=0pt,parsep=0pt]
    \item \textsf{Pointwise Verbalized Selection} \textbf{(PVS $without$ A):} Each passage $d_i \in \mathcal{D}$ is independently evaluated with a binary prompt for utility.
    \item \textsf{Listwise Verbalized Selection} \textbf{(LVS $without$ A):} The full candidate set $\mathcal{D}$ is presented, and the LLM selects the subset of useful passages.
    \item \textsf{Pointwise Verbalized Selection with pseudo-answer} \textbf{(PVS $with$ A):} Each passage is judged individually, with reference to a pseudo-answer.
    \item \textsf{Listwise Verbalized Selection with pseudo-answer} \textbf{(LVS $with$ A):} The entire set is presented, and the LLM selects useful passages with reference to a pseudo-answer.
\end{itemize}

\textbf{Utility-Based Ranking Methods}.  
These methods rank passages by their estimated utility. Approaches include:
\begin{itemize}[leftmargin=*,itemsep=0pt,topsep=0pt,parsep=0pt]
    \item \textsf{Listwise Verbalized Ranking \cite{zhang2024large}} \textbf{(LVR $without$ A):} The LLM produces an explicit ranking of passages, without a pseudo-answer.
    \item \textsf{Listwise Verbalized Ranking with pseudo-answer \cite{zhang2024large}} \textbf{(LVR $with$ A):} The LLM produces an explicit ranking with reference to a pseudo-answer.
    \item \textsf{Attention-based Utility Ranking \textbf{(AttR)} \cite{izacard2023atlas}:} Passage utility is inferred from the LLM’s attention weights during answer generation and normalized.
    \item \textsf{Likelihood-based Utility Ranking \textbf{(Likelihood)} \cite{shi2023replug}:} Utility is quantified by the conditional probability of generating the pseudo-answer given each passage, i.e., $P(a|q, d_i)$.
\end{itemize}

\vspace{-1mm}
\subsection{Experimental Results on \benchmark}
\heading{Utility-Based Selection Results}
Table \ref{tab:rq2_set} presents the utility-based selection performance (measured in F1 score) across various LLMs and datasets using different verbalized strategies.
Key observations are as follows:
(1) The listwise approach consistently outperforms the pointwise method across all LLMs, indicating its advantage in capturing contextual dependencies among passages for utility-based selection. 
Considering the accuracy when gold utilitarian  passages are empty, all methods have an over-selection problem, especially listwise methods. 
(3) Incorporating pseudo-answers during utility judgments generally leads to improved F1 performance. This finding is consistent with \citet{zhang2024large}. 

\heading{Utility-Based Ranking Results} 
Table \ref{tab:rq2_ranking} shows the utility-based ranking performance on different datasets. 
We can observe that: 
(1) The attention weights from the LLM during answer generation show the poorest alignment with utility, performing worse even than the standalone initial retrieval. This indicates that the internal attention mechanism is not a reliable indicator of a passage's contribution to the final answer.
(2) Both the likelihood and verbalized methods outperform the initial retrieval baseline across most datasets and models. This demonstrates that LLMs can effectively rank passages based on their utility for answer generation.
(3) The verbalized method with pseudo-answers achieves the best overall performance on most datasets. However, the likelihood method is superior on the TriviaQA dataset for all LLMs. As shown in Table \ref{tab:rq2_set}, LLMs can generate high-quality answers on TriviaQA, indicating that the queries in TriviaQA are simpler than those in other datasets. The superior performance of the likelihood method on this dataset suggests it is more sensitive to the quality of the pseudo-answer, benefiting more from accurate generations than the verbalized method. 

\heading{RAG Performance} 
To evaluate the answer generation performance of different utility-based passage selection methods, we employ the two top-performing approaches identified in Table \ref{tab:rq2_set}—verbalized pointwise and verbalized listwise selection, both incorporating pseudo-answers. 
The results are summarized in Table \ref{tab:rq2_set_rag}, from which we derive the following observations: 
(1) RAG using self-utility judgments generally outperforms the baseline of directly using the top-20 passages retrieved by BGE-M3. 
This trend holds across most datasets and LLMs, indicating the benefit of incorporating model-aware utility estimation in passage selection.
(2) Gold utilitarian passage sets are not transferable across different LLMs (Figure \ref{app:fig:top20_trasfer}); however, passages selected using listwise verbalized utility judgments produce RAG performance that generalizes well across models. In other words, LLMs do not necessarily achieve their optimal RAG performance when using passages chosen according to their own utility judgments. This effect is especially pronounced for weaker models, which often perform better when leveraging passages selected by stronger LLMs. These observations suggest that the utility captured by current selection methods tends to be general, rather than truly LLM-specific. 
Consequently, existing approaches are limited in their ability to identify passages with LLM-specific utility that can further enhance RAG performance. 
Future research should focus on developing more sophisticated strategies for selecting or generating passages that are genuinely tailored to the unique characteristics of each LLM, in order to fully realize the potential of retrieval-augmented generation.
\vspace{-1mm}
\section{Conclusion and Future Work}
In this work, we proposed and systematically studied the concept of \textbf{LLM-specific utility} for retrieval-augmented generation (RAG). Unlike traditional approaches that rely on static, human-annotated, or model-agnostic evidence, our results demonstrate that the passages most useful for one LLM are often not optimal for others, and that LLM-specific gold utilitarian passages consistently lead to better RAG performance than conventional human-annotated evidence. This finding fundamentally redefines RAG’s objective, highlighting the importance of tailoring evidence selection to the specific knowledge gaps and comprehension abilities of each model. 
To support this perspective, we introduced the LLM-specific utility judgment task and constructed a benchmark, i.e., \textbf{\benchmark}, spanning multiple LLMs and QA datasets. Our experiments show that while utility-aware selection methods—especially verbalized judgments with pseudo-answers—can improve RAG accuracy, most existing approaches still primarily capture a general notion of utility and fail to identify passages that maximize performance for individual LLMs. Interestingly, weaker models often perform better when using passages selected by stronger LLMs rather than their own selections, further emphasizing the need for generator-specific strategies. 
Overall, our work motivates a shift toward \textbf{LLM-centric evidence selection} in RAG. We hope our benchmark and findings inspire future research into utility estimation and retrieval methods that are genuinely tailored to the downstream LLM, ultimately improving the effectiveness and reliability of RAG systems.

For future work, several critical directions remain. First, although this analysis centers on single‑passage and context‑independent utility, exploring context‑dependent utility is an important next step. This includes investigating how the utility of a passage may change based on other passages already provided, particularly in multi‑hop settings, as well as incorporating broader utility factors such as factual grounding, and enhancing robustness. Second, extending the current binary, accuracy‑oriented framework to probabilistic utility and multi‑passage interactions will enable a more comprehensive understanding of LLM‑specific utility.  


\vspace{-1mm}
\begin{acks}
This work was funded by New Generation Artificial Intelligence-National Science and Technology Major Project of No. 2025ZD0123301, the National Natural Science Foundation of China (NSFC) under Grants No. 62302486 and No. 62441229, the Innovation Project of ICT CAS under Grants No. E561010 and E361140, and the Strategic Priority Research Program of the CAS under Grants No. XDB0680102. 
\end{acks}

\vspace{-1mm}
\section*{GenAI Usage Disclosure}
During the preparation of this work, the author(s) only used ChatGPT  or DeepSeek for writing and language polishing to improve the clarity, grammar, and readability of selected paragraphs. 
No other Generative AI tools were used in the research (including experimental design, data generation, or result interpretation) or in the writing of this paper. The author(s) remain solely responsible for all scientific and ethical aspects of this work. 
\bibliographystyle{ACM-Reference-Format}
\balance
\bibliography{reference}
\end{document}